\pdfoutput=1

\documentclass[11pt]{article}

\usepackage{EMNLP2022}

\usepackage{times}
\usepackage{latexsym}

\usepackage[T1]{fontenc}

\usepackage[utf8]{inputenc}

\usepackage{microtype}

\usepackage{inconsolata}

\usepackage{amsmath}
\usepackage{amssymb}
\usepackage{graphicx}
\usepackage{mathtools}
\usepackage{bbm}
\usepackage{multirow}
\usepackage{booktabs}
\usepackage{xspace}
\usepackage{algpseudocode}
\usepackage{algorithm}
\usepackage{eqparbox}
\usepackage{scalerel}

\usepackage{lipsum}
\usepackage[linecolor=orange,size=scriptsize]{todonotes}

\usepackage[normalem]{ulem}
\usepackage{comment}

\newcommand*\inlineimage[1]{\raisebox{-0.09\baselineskip}{\includegraphics[height=0.99\baselineskip]{#1}}}
\newcommand{\logo}{\inlineimage{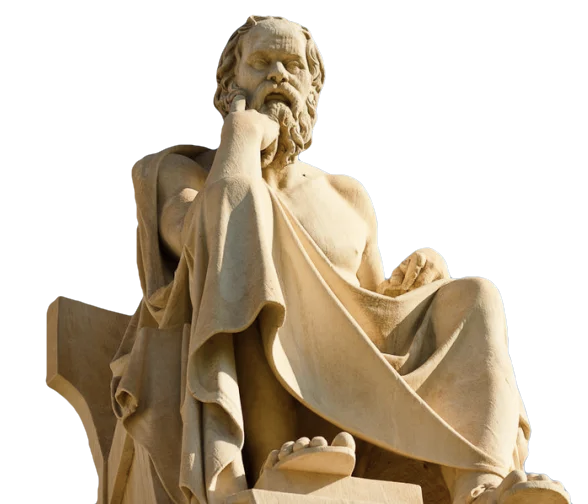}}

\title{\logo ~Maieutic Prompting: Logically Consistent Reasoning with\\ Recursive Explanations}

\author{Jaehun Jung\textsuperscript{$\dagger$} \hspace{.3cm} 
Lianhui Qin\textsuperscript{$\dagger$} \hspace{.3cm} 
Sean Welleck\textsuperscript{$\dagger\ddagger$} \hspace{.3cm}  \\
\textbf{\hspace{.3cm} Faeze Brahman}\textsuperscript{$\ddagger$}\hspace{.3cm}  
\textbf{\hspace{.3cm} Chandra Bhagavatula}\textsuperscript{$\ddagger$}  \hspace{.3cm} 
\textbf{\hspace{.3cm} Ronan Le Bras} \textsuperscript{$\ddagger$} \hspace{.3cm}
\textbf{\hspace{.3cm} Yejin Choi} \textsuperscript{$\dagger\ddagger$} \\
\textsuperscript{$\dagger$}Paul G. Allen School of Computer Science \& Engineering, University of Washington\\
\textsuperscript{$\ddagger$}Allen Institute for Artificial Intelligence \hspace{.3cm}  \\
\texttt{hoony123@cs.washington.edu}}

\newcommand{\method}[0]{\textsc{Maieutic prompting}\xspace}

\begin{document}
\maketitle
\begin{abstract}
Pre-trained language models (LMs) struggle with consistent reasoning; recently, prompting LMs to generate explanations that self-guide the inference has emerged as a promising direction to amend this. However, these approaches are fundamentally bounded by the correctness of explanations, which themselves are often noisy and inconsistent. In this work, we develop \method, which aims to infer a correct answer to a question even from the unreliable generations of LM. \method induces a tree of explanations \emph{abductively} (e.g. \textit{X is true, because \dots}) and \emph{recursively}, then frames the inference as a 
satisfiability problem over these explanations and their logical relations. We test \method for true/false QA on three challenging benchmarks that require complex commonsense reasoning. \method achieves up to 20\% better accuracy than state-of-the-art prompting methods, and as a fully unsupervised approach, performs competitively with supervised models. We also show that \method improves robustness in inference while providing interpretable rationales.\footnote{We share our code at \textit{https://github.com/jaehunjung1/ Maieutic-Prompting}.}

\end{abstract}

\section{Introduction}

Following the remarkable success of few-shot prompting over large language models (e.g. \citealp{gpt-3}), recent studies on prompting methods suggest that LMs' reasoning capability can be further promoted by generating a sequence of explanation for a given problem, prior to inferring the answer \cite{C-o-T, SC, GKP}. The so-called \textit{explanation-based prompting} helps an LM better elicit its knowledge and reason by leveraging its own generated explanations - whether it be commonsense knowledge \cite{GKP}, a solution for a math word problem \cite{C-o-T}, or the intermediate steps of program execution \cite{scratchpad}.

\begin{figure}[t]
\centering
\includegraphics[width=\linewidth]{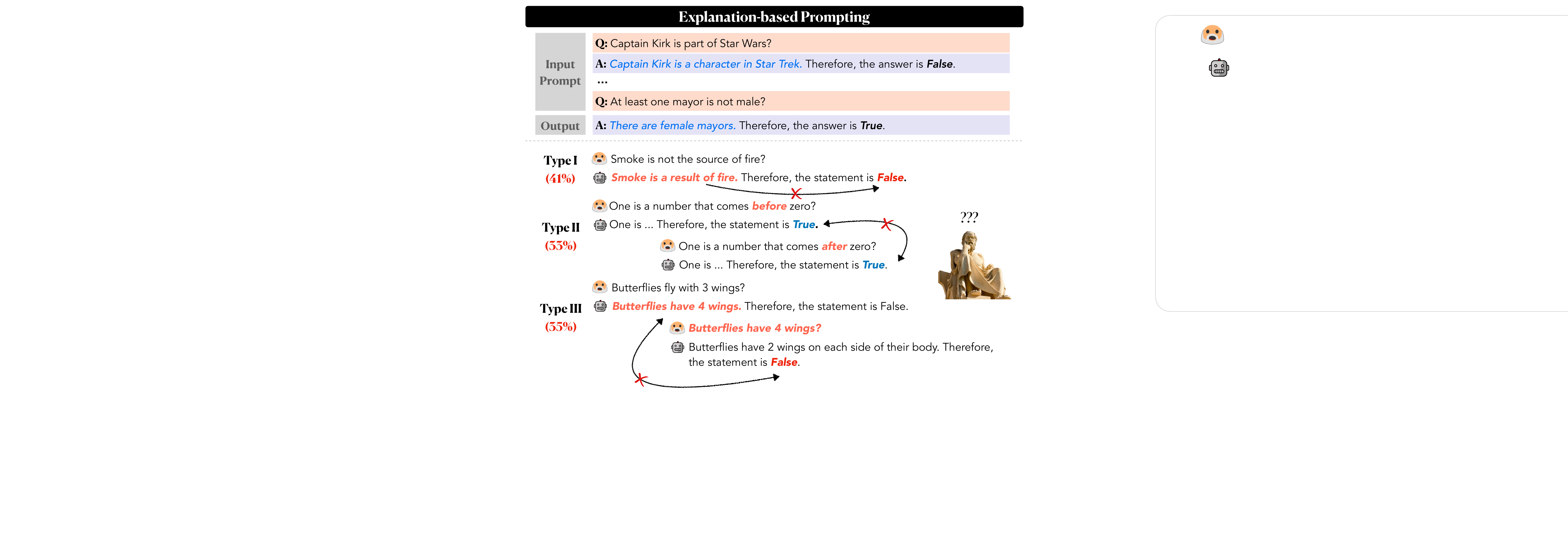}
\caption{Logical errors in explanation-based prompting: (1) explanation does not logically lead to the answer, (2) model is invariant to negation, and (3) falsifies its own explanation. We prompt 175B GPT-3 with 100 questions sampled from \citet{csqa2.0}.}
\vspace{-15pt}
\label{fig:motiv}
\end{figure}

\begin{figure*}[ht]
    \centering
    \includegraphics[width=\textwidth]{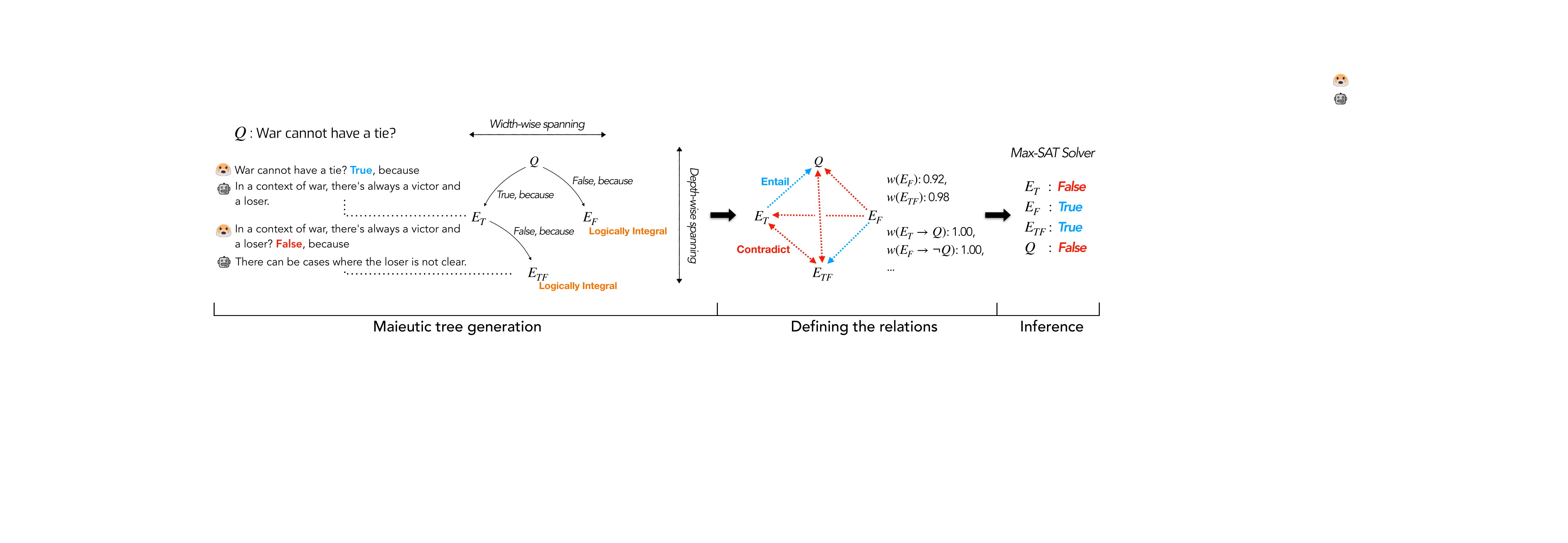}
    \vspace{-20pt}
    \caption{An overview of \method. Given a question $Q$, we generate \textit{maieutic tree} consisting of abductive and recursive explanations, define the relations between them, and employ MAX-SAT to find the best truth-value assignments to the explanations and $Q$.}
    \label{fig:overview}
    \vspace{-10pt}
\end{figure*}

Explanation-based prompting is intuitively motivated by the reasoning steps humans typically employ to solve a problem \cite{self-explaining}. However, we find that this intuition is faulty in practice, as model-generated explanations are often logically inconsistent and unreliable. For example, we manually inspected 100 samples from a QA task (Figure \ref{fig:motiv}) and found that for a considerable number of cases, (1) the explanation does not logically lead to the inferred answer, (2) the model infers the same label for a statement and its negation~\cite{kassner-schutze-2020-negated}, and (3) falsifies its own generated explanation. These findings raise fundamental questions on the role of explanations in LM inference: If the explanation is correct - is there a guarantee that the LM will infer a label consistent with the explanation? And if the explanation is wrong - is there a way to make use of even the wrong explanation in inferring the correct answer?

To this end, we propose \method, a novel few-shot inference method that infers a correct answer by enumerating a structure of explanations --- possibly noisy and contradictory --- and resolving them with a symbolic inference algorithm. Inspired by the maieutic method\footnote{\textit{Maieutic method brings out definitions implicit in the interlocutor's beliefs, ... is a method of hypothesis elimination, steadily identifying and eliminating those that lead to contradictions} \cite{maieutic}.} of Socrates, \method induces the LM to generate abductive explanations for diverse hypotheses with deep 
recursive reasoning, then collectively eliminates the contradicting candidates, resulting in consistent answers.

Figure \ref{fig:overview} shows the overview of \method. First, we prompt the LM to \textit{abductively} \cite{abductive} rationalize both possible answers, \textit{True} and \textit{False}, rather than generating a single explanation and then connecting it to one of the answer choices. Moreover, we do not expect the 1-hop explanations to be always correct; thus, we further validate the LM's confidence in its explanations by \textit{recursively} prompting the model with its own generation as the question. Our generation process derives a \textit{tree structure} of generated propositions, where one proposition establishes a logical ground for the correctness of one another.


To infer the answer for the original question, we quantify the strength of the LM’s \textit{belief} in each proposition and the \textit{logical relationships} between propositions in the maieutic tree. We then employ the weighted MAX-SAT \cite{Max-SAT} solver to \textit{collectively infer} the truth-values of all the propositions (including the original question) that best satisfy the set of observed relations. This way, we symbolically induce the subset of generations that makes the most probable and consistent inference. Our proposed method can run completely unsupervised with any few-shot promptable LM (e.g., GPT-3; \citealp{gpt-3}).

Our experiments show that the performance of \method exceeds that of all the few-shot prompting baselines (e.g., Chain of Thought; \citealp{C-o-T}) in three commonsense reasoning and fact verification benchmarks. \method performs up to 20\% better than other prompting methods, and performs on par or even better than supervised models. Further analyses show that \method is robust to perturbations in both the questions and prompts, and offers an interpretable interface to understand the rationale behind the model’s inference.

\section{Problem Setup and Background}
Our goal is to infer whether a given statement $Q$ makes sense, i.e. inferring the truth value $A$ of $Q$. Conventionally, this can be done through \textit{prompting} an LM with the following two methods:

\paragraph{Standard Prompting}
Let Q be a statement we want to infer the truth value of (i.e., either \textit{True} or \textit{False}). In standard few-shot prompting, the model-inferred answer $\hat{A}$ is defined as:
\begin{equation}
\small
    \hat{A} = \underset{A \in \{T, F\}}{\operatorname{argmax}}\, p_{LM}(A|Q, C),
\end{equation}
where $C = \{(q_1, a_1), \cdots, (q_k, a_k)\}$ denotes the $k$ examples for in-context learning.

\paragraph{Explanation-based Prompting}
In explanation-based prompting, the inference process is factorized into two steps:
\begin{equation}
\small
    \hat{A} = \underset{A \in \{T, F\}}{\operatorname{argmax}}\, \int_{E} p_{LM}(A|Q, E, C) \, p_{LM}(E|Q, C)
\end{equation}
Here, $E$ denotes the explanation generated prior to inferring the answer label, and $C = \{(q_1, e_1, a_1), \cdots, (q_k, e_k, a_k)\}$ includes $k$ examples of questions, explanations and answers. Since marginalizing over all $E$ is intractable, prior works resort to a sampling based approximation:
\begin{equation}
\small
\begin{split}
    & \hat{A} = \underset{A \in \{T, F\}}{\operatorname{argmax}}\, p_{LM}(A|Q, E, C),\\
    & \text{where}\,\, E \sim p_{LM}(E|Q, C)
\end{split}
\end{equation}

\section{Maieutic Prompting}
In this section, we introduce \method, which performs inference over a maieutic tree of generated explanations. First, we introduce \textit{logical integrity}, a key concept that is used to determine the reliability of propositions.

Language models often generate logically inconsistent propositions; for instance, in Figure \ref{fig:motiv}, the model infers \textit{True} when prompted with either \textit{``One is a number that comes before zero.''} or \textit{``One is a number that comes after zero.''}. In this sense, $p(\textit{True}|Q)$ does not provide a reliable value to determine whether $Q$ is true or not. We formalize this idea as \textit{logical integrity}: a proposition $Q$ is \textit{logically integral} when the LM consistently infers the truth value of $Q$ and $\neg Q$ (i.e. $Q$ as \textit{True} and $\neg Q$ as \textit{False}, or vice versa). Formally, we define a boolean function \texttt{integral}$(E)$ as follows:\footnote{Given $E$, $\neg E$ can be automatically generated simply by inserting a prefix (e.g. \textit{It is wrong to say that}), or prompting \textit{LM} to negate the given sentence.}
\begin{equation}
\small
\begin{split}
& \text{1.} \,\, \begin{aligned}[t] 
    & \underset{A \in \{T, F\}}{\operatorname{argmax}}\, p_{LM}(A|E, C) = T \,\, \text{and} \\
    & \underset{A \in \{T, F\}}{\operatorname{argmax}}\, p_{LM}(A|\neg E, C) = F \\
\end{aligned}\\
& \text{2.} \,\, \begin{aligned}[t] 
    & \underset{A \in \{T, F\}}{\operatorname{argmax}}\, p_{LM}(A|E, C) = F \,\, \text{and} \\
    & \underset{A \in \{T, F\}}{\operatorname{argmax}}\, p_{LM}(A|\neg E, C) = T \\
\end{aligned}\\[0.5em]
& \mathtt{integral}(E) = \mathbbm{1}_{\scaleto{\{\text{1 or 2 is satisfied\}}}{7pt}}.
\end{split}
\end{equation}
A statement is considered to be \textit{logically integral / True} when condition 1 is met, and \textit{logically integral / False} when condition 2 is met. Intuitively, the truth values of logically integral propositions are more credible than non-integral ones, to which LMs are inconsistent given a simple negation. For example, \textit{``One is a number that comes before zero.''} in Figure \ref{fig:motiv} would not be logically integral, as the model assigns same truth value to both $Q$ and $\neg Q$.

For the rest of section, we first search for logically integral propositions by constructing the maieutic tree (Section \ref{subsec:maieutic_tree_gen}), then quantify the relations between the propositions (Section \ref{subsec:relations}), based on which we infer the final answer (Section \ref{subsec:inference}).
\begin{figure*}[ht!]
    \centering
    \includegraphics[width=\textwidth]{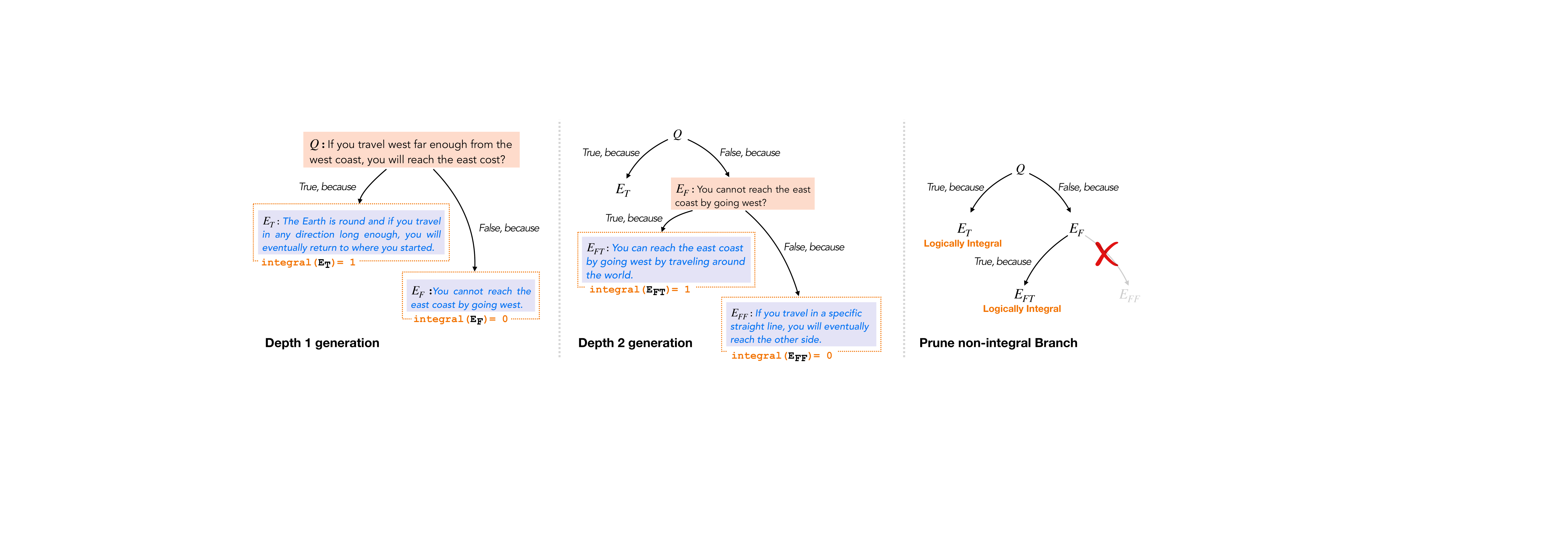}
    \vspace{-16pt}
    \caption{Illustrative example of \textit{maieutic tree} generation, with the max tree depth set to 2. For visual clarity, we generate only 1 $E_T$ and 1 $E_F$ per question and omit the width-wise spanning of knowledge.}
    \label{fig:tree_generation}
    \vspace{-10pt}
\end{figure*}

\subsection{Maieutic Tree Generation}
\label{subsec:maieutic_tree_gen}
\subsubsection{Abductive Explanation Generation}
\label{subsec: abd-gen}
Given a question, we require the LM to post-hoc rationalize both \textit{True} and \textit{False} labels. This abductive explanation generation has several advantages over an ad-hoc approach that first generates an explanation, then predicts the label. 
First, in the ad-hoc setting, the model is required to generate a discriminative explanation that helps in choosing one label over the other. Abductive generation \cite{abductive_commonsense_reasoning}, on the contrary, exposes the model to consider different possible answers rather than discriminating one, which often reveals an explanation that otherwise would not have been generated. 
Second, the label information would intuitively help LM elicit more specific explanations, mitigating the issue of a bland and generic generation which does not help the inference, a well-known weakness of LMs \cite{humanlike}.

Concretely, we define a function \texttt{abduction} which gets the statement $Q$ as the input and outputs a tuple of two abductive explanations with \textit{True}, \textit{False} given as the answer, respectively:
\begin{equation}
\small
\begin{split}
    & \mathtt{abduction}(Q) = (E_{T}, E_{F}) \\
    & \text{where} \,\, E_{A\in\{T, F\}} \sim p_{LM}(E|Q,A,C).
\end{split}
\end{equation}
Figure \ref{fig:overview} shows a concrete example of generating $E_T$ given $Q$. With $Q$, we prompt the model to rationalize \textit{True} as the answer: \textit{``War cannot have a tie? True, because''}, which then is completed by an explanation by LM \textit{``In a context of war, there's always a victor and a loser.''}.


\subsubsection{Depth-wise Knowledge Spanning}\label{subsec:depth}
As shown in Figure \ref{fig:motiv}, LM-generated explanations are noisy and inaccurate by nature. Prior works indirectly compensate for the untrustworthy generations by independently sampling multiple generations then aggregating them at the answer level (e.g. through majority voting; \citealp{SC}). Despite better performance, such an aggregation could still be brittle, as the inference fundamentally depends on the correctness of 1-hop explanations.

To enhance the robustness of reasoning, we hypothesize that the inference process should entail not only the \emph{breadth} of reasoning, but also the \emph{depth} of reasoning - whether the reasoning paths themselves are credible and consistent with each other. To do this, we require the LM itself to validate its own generations - by recursively prompting the LM with the generated explanations. As Figure \ref{fig:overview} shows, this corresponds to a depth-wise spanning of knowledge that induces a \textit{maieutic tree}, a multi-depth structure of generated propositions and relations between them.

Let $S_i$ denote the set of nodes at depth $i$ in the maieutic tree $\mathcal{T}$. Each node in $S_i$ is an explanation for an answer label (\textit{True} or \textit{False}), recursively generated given its parent node as the question:
\begin{equation}
\small
\begin{split}
    & S_i \subseteq \bigcup_{l \in \{T, F\}^{i-1}} \{E_{lT}, E_{lF}\}, \\
    & (E_{lT}, E_{lF}) = \mathtt{abduction}(E_l).
\end{split}
\end{equation}
Note that $\mathcal{T}$ is a full tree when the equality holds for all depths. For instance, in Figure \ref{fig:overview}, $E_{\textit{TF}}$ is generated by prompting the LM with its parent node $E_T$ and \textit{False}, i.e. $E_{TF} \sim p_{LM}(\cdot|E_T, \textit{F}, C)$. 

In practice, we sample multiple explanations with the same $Q$ and $A$ through nucleus sampling \cite{nucleus}. This corresponds to the width-wise spanning of knowledge, enhancing the diversity and coverage of generated explanations.

\subsubsection{When to Stop Generating}
\label{subsec:maieutic_tree}
Generating a full tree could be computationally expensive, as the number of generation grows exponentially with the maximum tree depth. Therefore, in each branch, we stop generating further once we reach a logically integral proposition; intuitively, this aligns with our goal to identify propositions that can be validated by the LM with confidence.

Figure \ref{fig:tree_generation} illustrates an example of maieutic tree generation where the maximum depth of the tree is set to 2. For visual clarity, we only generate one explanation per $Q$ and $A$. Given $Q$, we first generate $E_T$ and $E_F$, then validate whether each of them is logically integral. Since $E_T$ is logically integral, we stop generating in this branch, but continue generating from $E_F$ which is not logically integral. After reaching the maximum depth, we prune the branches leading to leaf nodes that are still not logically integral. This way, the final tree \emph{keeps only the generations that lead to a logically integral proposition.} We provide a formal description of the generation process in Appendix \ref{app:tree_generation}.

\subsection{Defining the Relations}\label{subsec:relations}
Now that we have generated the maieutic tree, we seek to define the relations between propositions and quantify their strength into scalar weights. For illustration, assume that an LM has generated the following $E_F$ for the given $Q$:
\begin{quote}
    Q: Captain Kirk is part of Star Wars?\\
    A: False, because \textbf{\textit{Captain Kirk is a character in Star Trek.}}
\end{quote}
The generation can be logically interpreted as follows: (1) the LM believes that \textit{Captain Kirk is a character in Star Trek}, (2) the LM believes that the proposition \textit{Captain Kirk is a character in Star Trek} can be a reason to deny that \textit{Captain Kirk is part of Star Wars}. Accordingly, we define \textit{belief} and \textit{consistency} to represent the two dimensions of the logical relationship.\\[0.5em]
\noindent
\textbf{Belief} $w_E$ corresponds to the LM's belief that the proposition $E$ is true (and therefore, $\neg E$ is false). To quantify \textit{belief}, we prompt the LM with $E$ and $\neg E$ respectively as a question, then comparing the probability assigned to \textit{True}:
\begin{equation}
\small
    w_E \coloneqq \frac{p_{LM}(T|E, C) - p_{LM}(T|\neg E, C)}{p_{LM}(T|E, C) + p_{LM}(T|\neg E, C)}.
\end{equation}
Note that calculating this does not require any additional prompting, as we already gained access to these values while checking for the logical integrity of each proposition.\\[0.5em]
\noindent
\textbf{Consistency} $w_{E, Q, A}$ corresponds to the consistency of the generated $E$ with the given $Q$ and $A$. Intuitively, if the LM is logically consistent, the likelihood of $E$ being generated given an answer (e.g., $E_F$ being generated given \textit{False}) should be larger than its likelihood given the opposite answer (e.g., $E_F$ being generated given \textit{True}). Following this intuition, we compute the consistency as:
\begin{equation}
\small
    w_{E,Q,A} \coloneqq \frac{p_{LM}(E|Q,A,C)}{p_{LM}(E|Q,A,C) + p_{LM}(E|Q,\neg A,C)}.
\end{equation}

\subsection{Inference}\label{subsec:inference}
The two types of relations formulate a set of unary and binary logical constraints, based on which we assign the truth values to all nodes in the maieutic tree $\mathcal{T}$, and in consequence, infer the answer to the original question. First, we represent $\mathcal{C}_{\textit{blf}}$ as the set of unary constraints. For each leaf node $E$ in $\mathcal{T}$,
\begin{equation}
\small
    c_{\textit{blf}} = \begin{cases}
        E & \text{if $E$ is \textit{logically integral / True}} \\
        \neg E & \text{if $E$ is \textit{logically integral / False}}.
    \end{cases}
\end{equation}
Note that all the leaf nodes in $\mathcal{T}$ are logically integral, hence we can count on the credibility of \emph{belief} for these nodes. We now define the set of all belief constraints $\mathcal{C}_{\textit{blf}}$ as:
\begin{equation}
\small
    \mathcal{C}_{\textit{blf}} = \{c_{\textit{blf}} \,\, \text{for}\,\, \forall E \in \text{leaf}(\mathcal{T})\}.
\end{equation}
For example, the nodes $E_F$ and $E_{TF}$ in Figure \ref{fig:overview} would have a belief constraint in $\mathcal{C}_{\textit{blf}}$.

Likewise, for \emph{consistency}, we define $\mathcal{C}_{con}$ as the set of binary constraints using logical implication. For each edge $(E_{l}, E_{lA})$ in $\mathcal{T}$,
\begin{equation}
\small
\begin{split}
    &c_{\textit{con}} = \begin{cases}
        E_{lA} \rightarrow E_{l} & \text{if $A$ = \textit{True}} \\
        E_{lA} \rightarrow \neg E_{l} & \text{if $A$ = \textit{False}}
    \end{cases}\\
    & \mathcal{C}_{\textit{con}} = \{c_{\textit{con}} \,\, \text{for}\,\, \forall (E_{l}, E_{lA}) \in \text{edge}(\mathcal{T})\}.
\end{split}
\end{equation}
Our objective is to assign the truth values for all $E$s and the root node $Q$ in $\mathcal{T}$, such that we maximize
\begin{equation}
\label{eqn:obj}
\small
\begin{split}
        \sum_{\scaleto{c \in \mathcal{C}_{\textit{blf}} \cup \mathcal{C}_{\textit{con}}}{7pt}} w_c \cdot \mathbbm{1}_{\scaleto{\{c = \text{True}\}}{7pt}},
\end{split}
\end{equation}
which sums up the weights of satisfied constraints.

This problem is naturally formulated as weighted MAX-SAT, which is a problem of determining truth values of variables that maximize the weight of satisfied clauses. The problem can be algorithmically solved using an off-the-shelf solver. 

\vspace{-10pt}
\begin{table*}[t]\centering
    \resizebox{.9\textwidth}{!}{
    \begin{tabular}{ llcccccccc }\toprule
        \multicolumn{2}{c}{\textbf{Dataset}} & \multicolumn{3}{c}{\textbf{Com2Sense}} & \multicolumn{2}{c}{\textbf{CSQA 2.0}} & \multicolumn{3}{c}{\textbf{CREAK}} \\\cmidrule(lr){3-5}\cmidrule(lr){6-7}\cmidrule(lr){8-10}
        \multicolumn{2}{c}{\textbf{Model}} & \textit{dev} & \textit{test} & \textit{pairwise} & \textit{dev} & \textit{test} & \textit{dev} & \textit{test} & \textit{contrast}\\\midrule
        \multirow{6}{*}{\rotatebox[origin=c]{90}{\parbox[c]{1.9cm}{\centering Supervised}}} 
        & RoBERTa-large~\cite{roberta} & 62.8 & 59.4 & 33.3 & - & - & 80.6 & 80.3 & 61.5 \\
        & T5-large~\cite{t5} & 62.8 & 60.6 & 41.8 & 53.8 & 54.6 & - & - & - \\
        & T5-3B~\cite{t5} & 73.2 & - & - & - & 60.2 & 85.6 & 85.1 & 70.0 \\
        & UnifiedQA-3B~\cite{unifiedqa} & 75.1 & \textbf{71.3} & \textbf{51.3} & - & - & - & - & - \\
        & T5-11B~\cite{t5} & \textbf{77.2} & - & - & 68.5 & 67.8 & \textbf{89.5} & - & \textbf{75.2} \\
        & Unicorn-11B~\cite{unicorn} & - & - & - & \textbf{69.9} & \textbf{70.2} & - & - & - \\\midrule
       \multirow{5}{*}{\rotatebox[origin=c]{90}{\parbox[c]{1.9cm}{\centering Prompting}}} 
        & Standard & 58.1 & - & - & 54.1 & - & 60.3 & - & 55.2 \\
        & Chain of Thought~\cite{C-o-T} & 61.6 & - & - & 59.6 & - & 64.8 & - & 59.4 \\
        & Self Consistency~\cite{SC} & 61.4 & - & - & 60.8 & - & 70.5 & - & 64.8 \\
        & GKP~\cite{GKP} & 61.8 & - & - & 59.7 & - & 75.4 & - & 68.2 \\
        & \method (Ours) & \textbf{72.5} & \textbf{75.0} & \textbf{68.7} & \textbf{69.5} & \textbf{68.3} & \textbf{85.2} & \textbf{85.3} & \textbf{77.4} \\\bottomrule
    \end{tabular}
    }
    \vspace{-6pt}
    \caption{Experimental results of \method and baseline methods on three benchmark datasets. We differentiate supervised baselines (upper section) from prompting methods (lower section), and bold the best numbers for each section. \method with GPT-3 outperforms all prompting baselines with the same model, while being competitive against billion-scale supervised LMs. }
    \label{tab:performance}
\end{table*}

\vspace{5pt}
\subsection{Verifier Model}
One limitation of the consistency definition in Section \ref{subsec:relations} is that it only considers the relationship between a parent node and a child node. Since the definition builds upon the likelihood of each generation from an LM, we cannot take into account the relationships across branches, e.g. $E_T$ and $E_F$ in Figure \ref{fig:tree_generation}. This motivates us to introduce a small NLI model as a verifier, which can infer the relationship between an arbitrary pair of nodes in $\mathcal{T}$.
Following previous works \cite{minervini_nli, swap_nli}, we convert the NLI labels into logical relations as following:
\begin{equation}
\small
\begin{split}
\begin{gathered}
    \texttt{Entail}(E_1, E_2): E_1 \rightarrow E_2 \\ \texttt{Contradict}(E_1, E_2): E_1 \rightarrow \neg E_2.
    \end{gathered}
\end{split}
\end{equation}
For all pairs of nodes $(E_1, E_2) \in \text{node}(\mathcal{T})^2$, $E_1 \neq E_2$, we obtain either $E_1 \rightarrow E_2$ or $E_1 \rightarrow \neg E_2$ if $E_1$ entails or contradicts $E_2$. For NLI-based clauses, we fix the weights to 1.\footnote{We also tried using the label probability assigned by NLI model as weight, but fixing it to 1 yielded better results.} While the objective function (Eq. \ref{eqn:obj}) stays the same, $\mathcal{C}_{\textit{con}}$ is now replaced with $\mathcal{C}_{\textit{NLI}}$, a set of clauses induced by the verifier model.

\section{Experiments}
\paragraph{Datasets}
We evaluate \method on three commonsense reasoning and fact verification benchmarks in binary QA format: Com2Sense \cite{com2sense}, CSQA 2.0 \cite{csqa2.0}, CREAK \cite{creak}.
Despite the simple format, these datasets require a substantial amount of knowledge and robust reasoning, making them challenging even for the billion-scale fine-tuned LMs (Table \ref{tab:performance}).

\paragraph{Baselines}
We compare our method with both the few-shot prompting methods and supervised models. Along with the standard prompting, we include Chain of Thought \cite{C-o-T}, Self-Consistency \cite{SC} and Generated Knowledge Prompting (GKP) \cite{GKP}. For supervised models, we consider the strong baselines used for the respective dataset, such as T5 \cite{t5}, UnifiedQA \cite{unifiedqa} and Unicorn \cite{unicorn}.

\paragraph{Configuration Details}\label{sec:config} For all prompting methods, we use the same set of 6 demonstration examples and the same version of GPT-3 (\textit{text-davinci-001}) as the LM. We determine the hyperparameters of \method and baselines based on the dev set performance on the benchmarks. In maieutic tree generation, we set the maximum depth to 2. For depth 1, we use nucleus sampling ($p = 1.0$) \cite{nucleus} to generate 3 $E_T$s and 3 $E_F$s from $Q$. For depth 2, we use greedy decoding to generate 1 $E_T$ and 1 $E_F$ from each parent node. This constrains the generated tree to have at most 18 nodes excluding the original $Q$.\footnote{Both GKP and Self Consistency employ an ensemble strategy, generating $N$ different samples of explanations then aggregating their answers. For a fair comparison with ours, we set $N=20$ for both methods, generating more explanations than the maximal possible size of the maieutic tree.} In Section~\ref{subsec:ablation}, we conduct an ablation study on this depth-adaptive decoding scheme and analyze the effect of the tree size. For the main experiments, we use RoBERTa~\cite{roberta} fine-tuned on MNLI \cite{mnli} as a verifier with 90.2\% accuracy on MNLI dev set, and RC2 \cite{rc2} as a MAX-SAT solver.

\subsection{Benchmark Performance}
Table \ref{tab:performance} presents overall evaluation results of \method along with the prompting and supervised baselines. \method significantly outperforms all prompting methods across all benchmarks. Notably, GKP and Self Consistency ensembled more 1-hop explanations than the maximal size of the maieutic tree; our superior performance compared to these methods confirms the sample efficiency of depth-wise knowledge spanning. Moreover, \method is the only prompting method that performs better than even the smallest supervised baseline (RoBERTa-large) in Com2Sense and CREAK. In fact, \method allows us to use an off-the-shelf LM to achieve comparable performance to a large \textit{fine-tuned} LM by simply plugging in our inference algorithm. In Appendix \ref{app:multi_hop} we also provide experiments on StrategyQA \cite{strategyqa}, to evaluate the generalizability of \method in multi-hop setting.


\begin{figure}
    \centering
    \includegraphics[width=.49\textwidth]{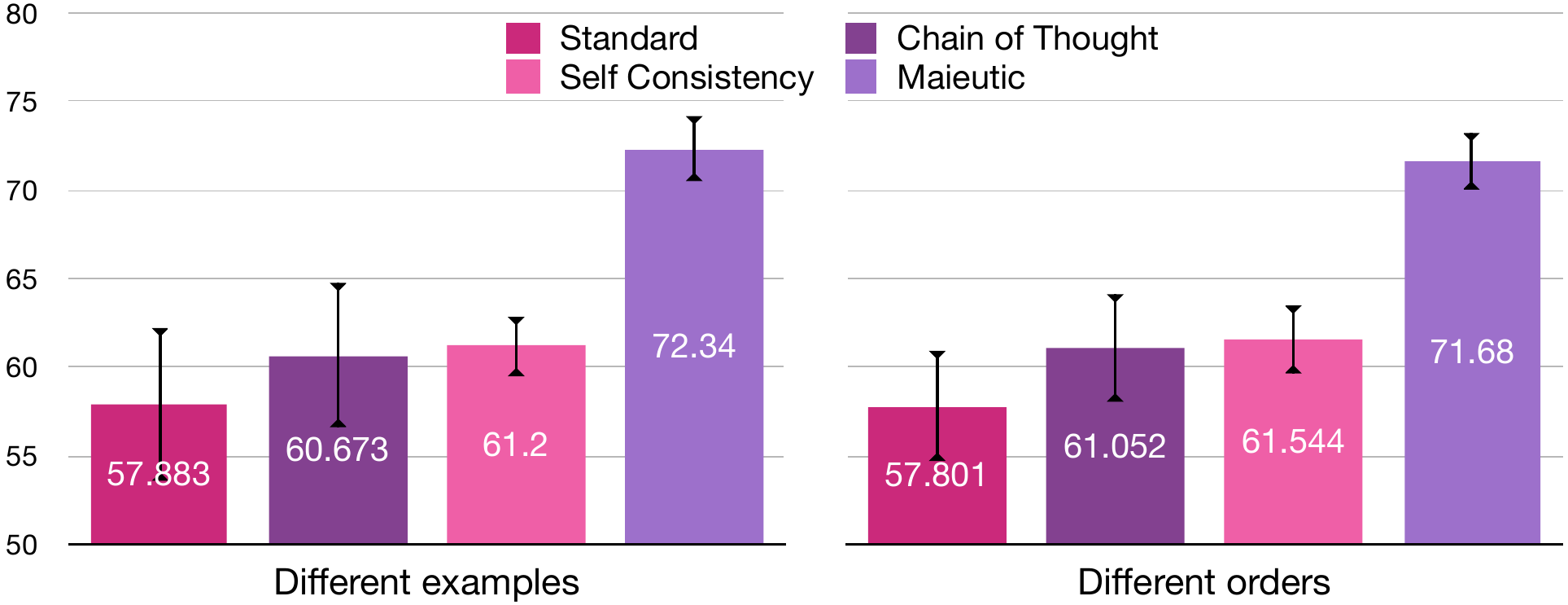}
    \vspace{-16pt}
    \caption{Robustness of prompting methods under different few-shot examples / different order of examples. We compare the mean and standard deviation of Com2Sense dev set accuracy. }
    \label{fig:robust_prompt}
    \vspace{-10pt}
\end{figure}

\subsection{Robustness Analysis}
We perform additional analyses to understand the working of our method under semantic perturbations and different prompt formats.

\paragraph{Robustness to semantic perturbations}
In addition to the standard accuracy, we report two additional metrics called \textit{pairwise accuracy} and \textit{contrast set accuracy} in Table \ref{tab:performance}. In Com2Sense test set and CREAK contrast set, each question is paired with its complimentary counterpart, of which the surface form is similar but the answer should be the opposite (e.g. \textit{``Barack Obama has daughters.'' vs ``Barack Obama has no daughter.''}), testing the models' robustness to semantic perturbations. In these metrics, the gap between \method and baselines widens substantially, indicating the robustness of our method against semantic perturbations.

\paragraph{Robustness to different prompts}
Prior works revealed that prompting performance could be sensitive to few-shot examples and their order \cite{order, calibrate}. We investigate whether this holds true for \method, as shown in Figure \ref{fig:robust_prompt}. We compare different prompting methods run with 3 different sets of few-shot examples (left), and 5 different permutations of the few-shot examples (right). In both settings, while Self Consistency and \method are much more stable then the other two, our method has slightly less variance.

\begin{table}[t]
    \centering
    \small
    \begin{tabular}{lc}\toprule
        \textbf{Model} & \textbf{Accuracy} \\\midrule
        Non-abductive generation & 68.4 \\
        All greedy decoding (no depth-adaptive) & 67.2 \\
        All nucleus sampling (no depth-adaptive) & 72.0  \\
        Likelihood-based consistency & 65.6 \\ \midrule
        Maieutic Prompting & \textbf{72.5} \\\bottomrule
    \end{tabular}
        \vspace{-6pt}
    \caption{Ablation study on Com2Sense Dev set. The best configuration is with abductive generation, depth-adaptive decoding and verifier-based consistency.}
    \label{tab:ablation}
\end{table} 
\begin{table}[t]
    \centering
    \small
    \begin{tabular}{lccccc}\toprule
        \textbf{Dimension} & \textbf{1} & \textbf{2} & \textbf{3} & \textbf{5} & \textbf{10}\\\midrule
        Depth & 61.3 & 72.5 & 72.4 & - & - \\
        Width & 62.4 & 66.5 & 72.5 & 71.5 & 72.1 \\\bottomrule
    \end{tabular}
        \vspace{-6pt}
    \caption{Performance of \method on Com2Sense with different maieutic tree sizes.}
    \label{tab:tree_size}
    \vspace{-10pt}
\end{table}

\subsection{Ablation Study}
\label{subsec:ablation}

We ablate different components of \method to investigate their respective contributions as shown in Table \ref{tab:ablation}.
\paragraph{Generation} First, we consider \method without abductive generation --- we generate each explanation without providing an answer label, i.e. in an identical fashion to Chain of Thought. In this setting, the performance of \method degrades by 4\%, alluding to the importance of abductive generation in eliciting the latent knowledge from LM. Next, we ablate the depth-adaptive decoding mechanism (Section \ref{sec:config}), by applying either greedy decoding or nucleus sampling for all depths of the maieutic tree. \textit{All greedy decoding} restrains width-wise spanning of knowledge, hence leads to large degradation of performance. \textit{All nucleus sampling} performs much more comparably with our best configuration, although the stochastic decoding produces slightly more errors in the explanations.

\begin{figure}
    \centering
    \includegraphics[width=.48\textwidth]{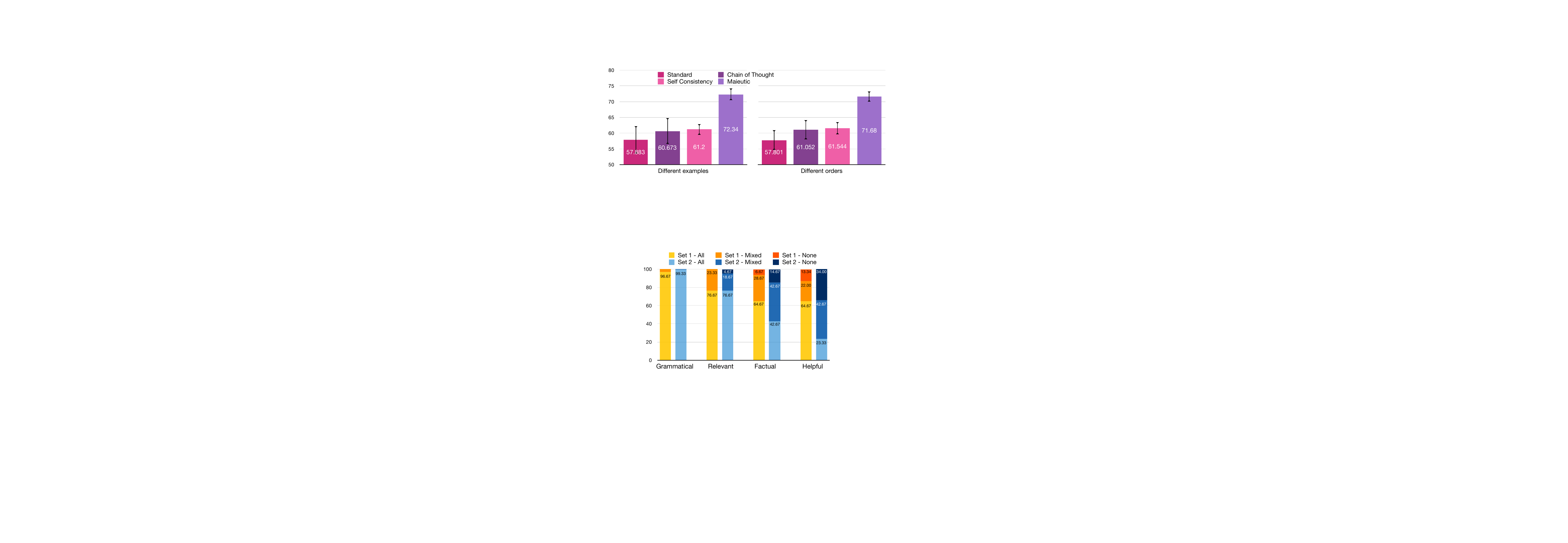}
    \vspace{-18pt}
    \caption{Human evaluation results (\citeauthor{krippendorff}’s alpha = 0.64; substantial inter-annotator agreement). To minimize subjectivity, we use a strict 3-level scale, where annotators choose \textit{All} only when all the statements in the \textit{true Es} are desirable (e.g. grammatical) on its own, \textit{Mixed} when at least one $E$ is undesirable, and \textit{None} otherwise.}
    \label{fig:human_eval}
    \vspace{-10pt}
\end{figure}

\begin{figure*}[ht]
    \centering
    \includegraphics[width=\textwidth]{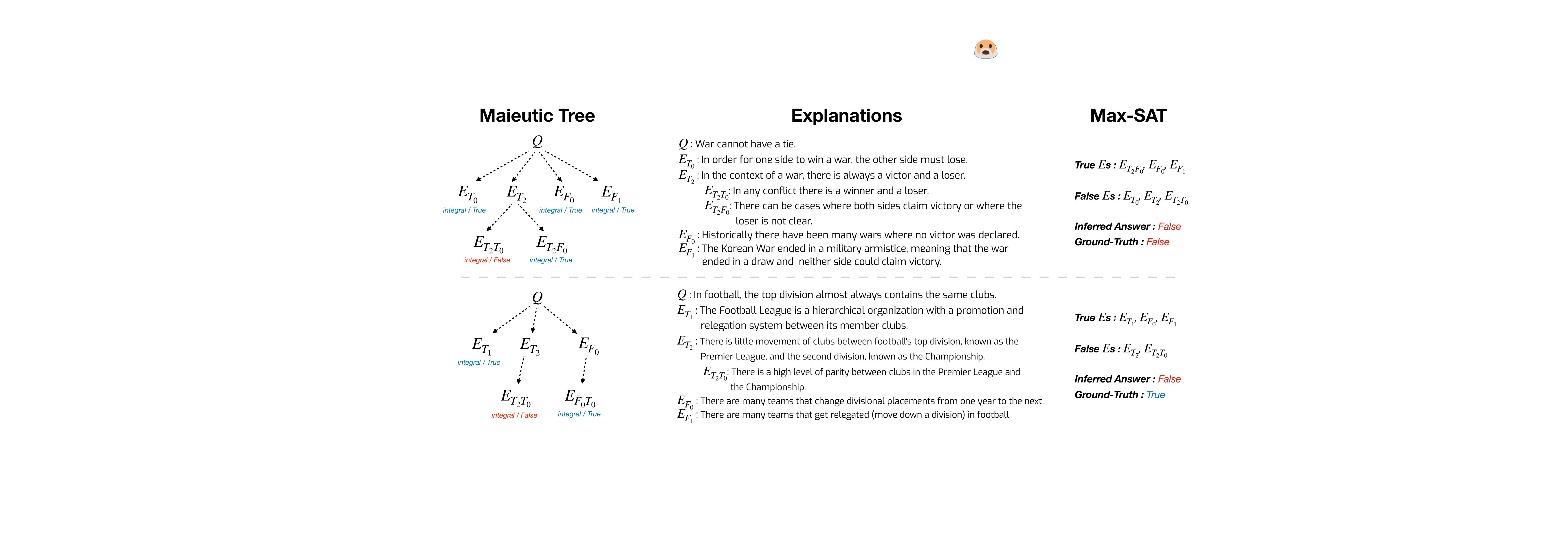}
    \vspace{-16pt}
    \caption{Examples of \method, each with correct and wrong answer. Even in the latter case, the generated explanations make sense toward the inferred answer. We provide more examples in Appendix \ref{app:examples}. }
    \label{fig:case_study}
    \vspace{-10pt}
\end{figure*}

\paragraph{Consistency} We ablate the NLI-based clauses and replace them with the original $C_{con}$ discussed in Section \ref{subsec:relations}. With the likelihood-based $C_{con}$, the accuracy reduces by about 7\%, but still prevails over the prompting baselines in Table \ref{tab:performance}. The verifier model indeed benefits the inference process by providing more accurate relations between generated explanations, although our method performs competently even without the access to the verifier.

\paragraph{Effect of tree size}
We also investigate how the size of the maieutic tree influences the performance. In Table \ref{tab:tree_size}, we present the performance of \method on Com2Sense dev set with various values of maximal depth and width. In both dimensions, the accuracy saturates after a certain threshold. We attribute this to (1) the topic drift in generation which intensifies as the depth grows, (2) larger overlaps in generated knowledge as we sample more explanations width-wise.

\subsection{Human Evaluation}
We qualitatively analyze actual inference results of \method through human evaluation. For each sample, we first retrieve \textit{true $E$s} (the set of generated $E$s that are inferred to be \textit{True} by \method), then evaluate them over the four criteria from \citet{GKP}: 
(1) \textit{Grammaticality} of the explanations, (2) \textit{Relevance} of the explanations to the question, (3) \textit{Factuality}: whether the explanations states facts, and (4) \textit{Helpfulness}: whether the explanation explicitly leads to the correct answer. Six NLP experts scored 100 examples sampled from CSQA 2.0 dev set, of which 50 were answered correctly (Set 1) and 50 were answered wrongly by the model (Set 2).

Figure \ref{fig:human_eval} presents the evaluation results. For both sets, over 99\% of the \textit{true $E$s} are grammatically perfect, and most of them provide relevant evidence to the question.\footnote{It is natural that some of the true $E$s are not directly relevant to $Q$, but still contribute to the inference by validating other $E$s.} Surprisingly, the LM often generates both factual and helpful explanations even when its answer is different from the ground truth: 42\% of the \textit{true $E$s} for incorrectly answered examples are perfectly factual, and 23\% of them are completely helpful in correctly answering the question. We find that in many of these cases, the questions did not have a clear-cut answer; as exemplified in Figure \ref{fig:case_study}, the explanations generated and validated by \method are compelling enough as an alternative to the ground-truth answer.

\section{Related Work}
Prior works have leveraged natural language explanations (NLEs) to promote model reasoning, either by training a model to explain \cite{cage, e-snli, chen2022can, wiegreffe2021teach}, or generating answers to templated queries and distantly supervised rationales \cite{self-talk, DBLP:conf/aaai/BrahmanSRC21}. Incorporated with in-context learning (\citealp{gpt-3}; \textit{inter alia}), these efforts have led to explanation-based prompting \cite{C-o-T, SC, GKP, lampinen2022can}. Other works aim to improve model interpretability with NLEs, training a model that explains its inference post-hoc or in parallel with the answer \cite{e-snli, wt5, jacovi2021contrastive}. Unlike these works, the explanations in our work are designed to be intrinsic \cite{iml}; the explanations themselves explicitly participate in the inference.

Meanwhile, recent observations reveal that LM explanations are unreliable, as they often lack logical consistency and are not factually grounded \cite{unreliability, kassner-schutze-2020-negated}. This is in part due to the broader limitations of generative LMs, which assign high probability to unlikely sentences \cite{welleck2020consistency, surfaceform} and are sensitive to semantic perturbations \cite{elazar}. \method overcomes these limitations by avoiding the use of explanations ``as-is'', and modeling the relationships between explanations.

Another line of works apply symbolic methods on top of LMs to improve their consistency, spanning from a lexical constraint on sequence decoding \cite{neurologic-decoding} to a symbolic world model \cite{system2} and discrete operations \cite{nerd, openai-math}. Other works explore how to train a model that simulates the symbolic reasoning process, such as logical transformation \cite{flexible} and consistent generation of beliefs \cite{beliefbank, towards}. However, these models require a curated set of human annotations that limits their application to specific domains. \method generalizes these neuro-symbolic approaches in an unsupervised setup, employing MAX-SAT algorithm to symbolically determine the true subset from a noisy pool of neural generations.

\section{Conclusion}
In this work, we propose \method, a novel few-shot inference method inspired by the Socratic way of conversation. We systematically generate a tree of explanations that bear logical relations between each other, then find the truth values that max-satisfy these relations. Empirical results show that \method is both competitive and robust compared to diverse baselines, while providing intrinsic interpretations over its inference.

\section*{Limitations}
\paragraph{Extension to different task formats} In this work, we limit our experiments to  validating a given statement. In future works, we aim to extend our method over a broader range of tasks, e.g. multiple-choice QA. A potential strategy could be binarizing multiple-choice options to respective statements and scoring them with \method, e.g. using the sum of weight of satisfied clauses from MAX-SAT.

\paragraph{Modeling relationships between trees} \method models the relations between the nodes in each maieutic tree to infer a consistent answer. The scope of modeled relationships, however, could be further generalized beyond a single tree - a span of knowledge generated for one question could serve as the evidence for another question. Indeed, modeling the relationship between questions is an active area of research \cite{datapoints}. We envision that the knowledge elicited from \method could further be enriched through this type of generalization.

\section*{Acknowledgements}
This work was funded in part by the Natural Sciences and Engineering Research Council of Canada (NSERC) (funding reference number 401233309), DARPA MCS program through NIWC Pacific (N66001-19-2-4031), and the Allen Institute for AI. We also thank OpenAI for providing access to the GPT-3 API.

\bibliography{anthology,custom}
\bibliographystyle{acl_natbib}

\newpage
\onecolumn
\appendix

\section{Tree Generation Algorithm}\label{app:tree_generation}
\begin{algorithm}[H]
\caption{Maieutic tree generation}\label{alg:tree_generation}
\begin{algorithmic}\small
\Require Question $Q$, Max tree depth $D$
\Ensure Maieutic tree $\mathcal{T}$\\
$\mathcal{T} \gets \mathtt{init}(Q)$ \Comment{initialize the tree with $Q$}
\For{$d \in \{1, \cdots, D\}$} \Comment{generate nodes}
    \State $S_i \gets \emptyset$
    \For{$E \in S_{i-1}$}
        \If{$\mathtt{integral}(E) = 1$}
            \State $S_i \gets S_i \cup \mathtt{abductive}(E)$
        \EndIf
    \EndFor
    \State $\mathcal{T}\mathtt{.add}(S_i)$
\EndFor\\
$V \gets \{E\text{; }\mathtt{integral}(E) = 0\,\,\text{for all } E \in \text{leaf}(\mathcal{T})\}$ \Comment{set of non-integral leaf nodes}
\While{$V \neq \emptyset$} 
\State $\mathcal{T}\mathtt{.remove}(V)$ \Comment{prune the non-integral leaf nodes}
\State $V \gets \{E\text{; }\mathtt{integral}(E) = 0\,\,\text{for all } E \in \text{leaf}(\mathcal{T})\}$
\EndWhile
\end{algorithmic}
\end{algorithm}

\section{Dataset Details}\label{app:dataset_statistics}
\begin{table}[htb]
    \centering
    \small
    \begin{tabular}{cccc}\toprule
        \textbf{Dataset} & \textbf{Com2Sense} & \textbf{CSQA 2.0} & \textbf{CREAK} \\\midrule
        Train / Dev / Test split size & 804 / 402 / 2779 & 9282 / 2544 / 2517 & 10176 / 1371 / 1371 \\
        Average \# of tokens & 21 & 11.3 (words) & 10.8 \\\bottomrule
    \end{tabular}
    \caption{We evaluate \method in three commonsense reasoning and fact verification benchmarks - Com2Sense, CSQA 2.0 and CREAK. Com2Sense and CSQA 2.0 consist of adversarial commonsense questions generated to mislead a proxy model. CREAK tests for a combination of commonsense reasoning and accurate fact retrieval, consisting of long-tail questions such as \textit{``Harry Potter can teach how to fly on a broomstick?''}. Table \ref{tab:dataset_statistics} presents key statistics of the three datasets.}
    \label{tab:dataset_statistics}
\end{table}

\section{Multi-hop Reasoning on StrategyQA}\label{app:multi_hop}
\begin{table}[htb]
    \centering
    \small
    \begin{tabular}{cccccc}\toprule
        \textbf{Model} & \textbf{Standard} & \textbf{C-o-T} & \textbf{Maieutic} & \textbf{C-o-T (Multi-hop)} & \textbf{Maieutic (Multi-hop)} \\\midrule
        Accuracy & 56.3  & 58.2  & 60.7  &  57.9 & 61.4 \\\bottomrule
    \end{tabular}
    \caption{Results on StrategyQA}
    \label{tab:multi_hop}
\end{table}
To further evaluate the generalizability of \method, we conduct additional experiments on multi-hop reasoning over StrategyQA \cite{strategyqa} dev split. Note that the original evaluation setting for StrategyQA presupposes access to Wikipedia articles, from which the gold knowledge could be retrieved from; hence the benchmark as-is may not represent the best evaluation setting for few-shot prompting methods.

To better address the multi-hop nature of the dataset, we add a straightforward adjustment to both C-o-T and Maieutic Prompting, to first decompose the original question into 2-3 minor questions and then generate the explanation and answer. We denote this as \textbf{Multi-hop} in Table \ref{tab:multi_hop}.

Consistent with the original experimental results, Maieutic Prompting yields promising improvement compared to both the standard / C-o-T prompting. The result attests to both the generalizability of Maieutic Prompting to multi-hop setting and the importance of reasoning algorithm in challenging scenarios.

\newpage

\section{Inference Examples}\label{app:examples}
\vspace{1cm}
\begin{figure*}[!htb]
    \centering
    \includegraphics[width=\textwidth]{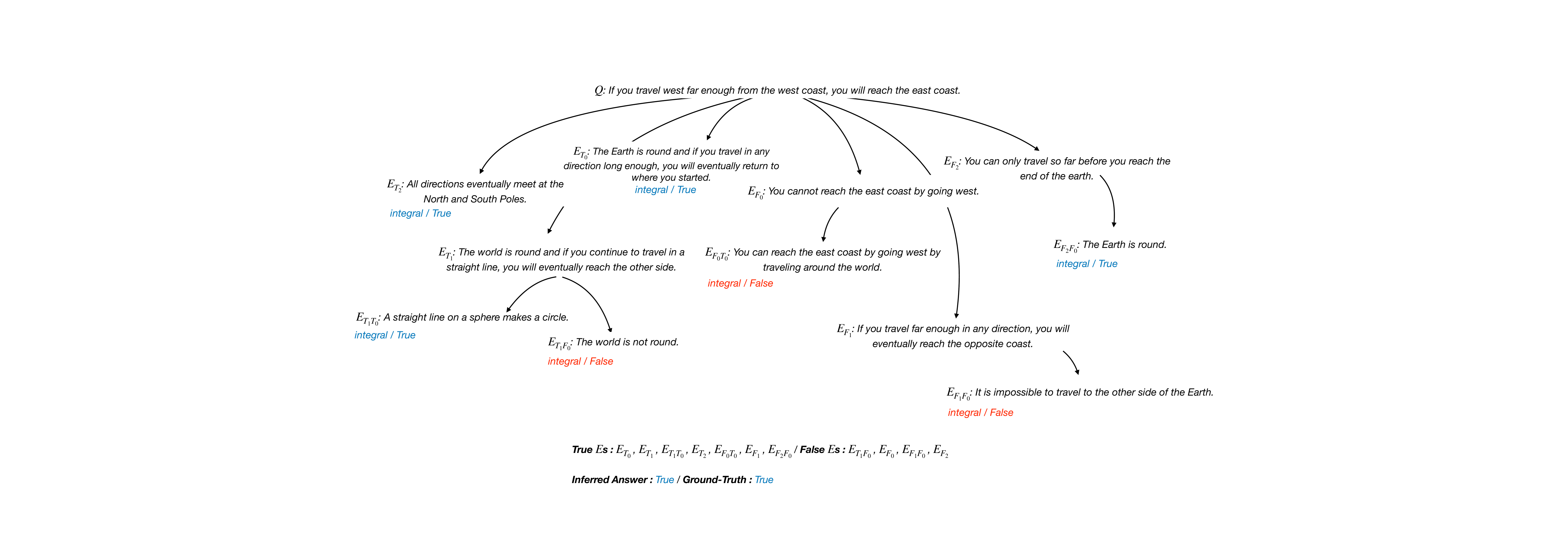}
    \caption{Example of correct inference by \method. We show the generated maieutic tree along with the assigned truth-values to each propositions.}
\end{figure*}
\vspace{2cm}
\begin{figure*}[!htb]
    \centering
    \includegraphics[width=\textwidth]{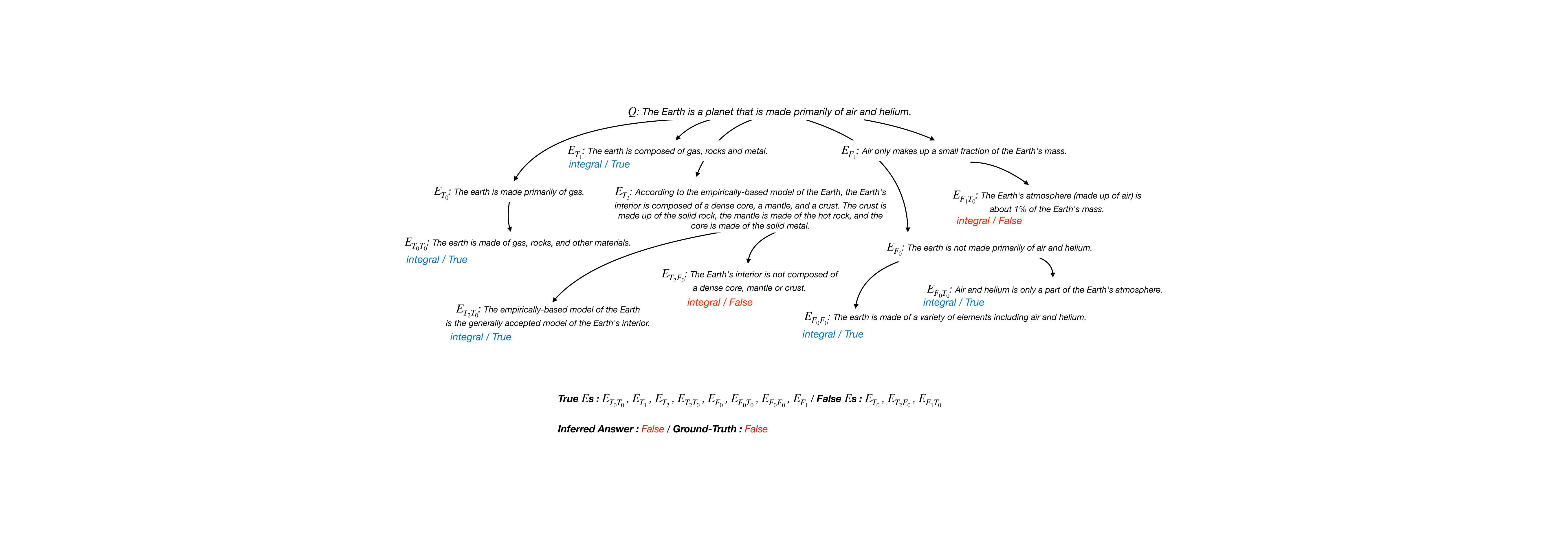}
    \caption{(continued) Example of correct inference by \method.}
\end{figure*}

\begin{figure*}[!htb]
    \centering
    \includegraphics[width=\textwidth]{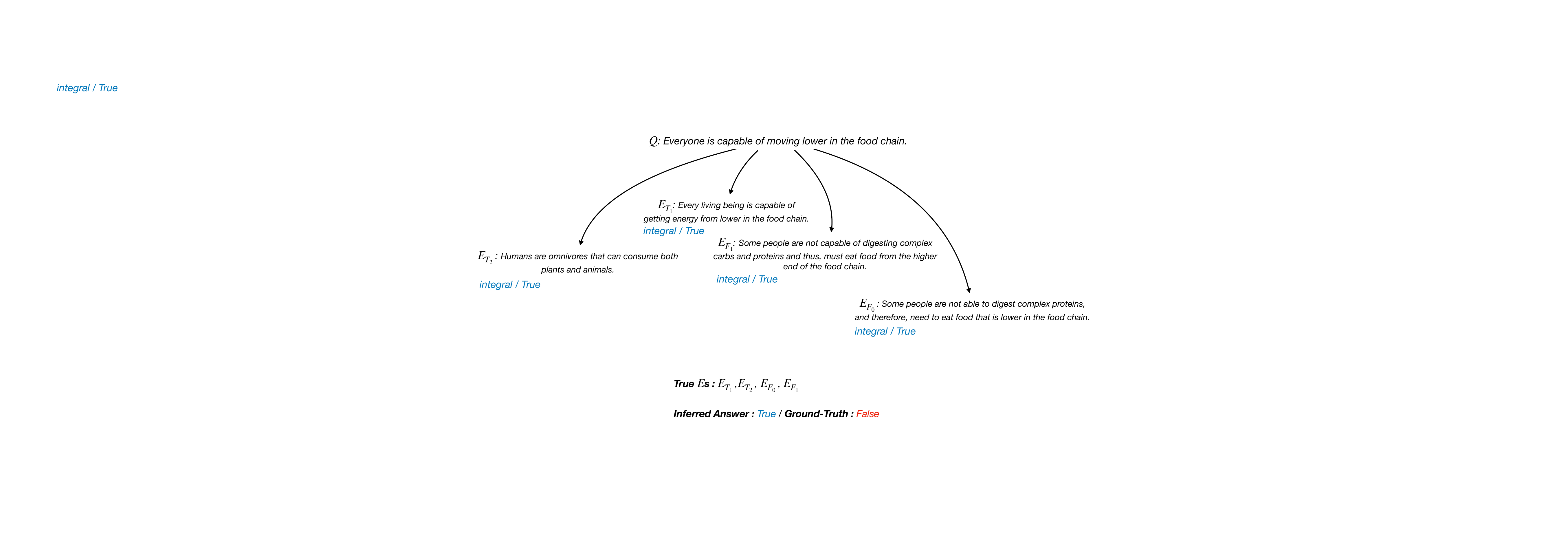}
    \caption{Example of incorrect inference by \method.}
\end{figure*}

\begin{figure*}[!htb]
    \centering
    \includegraphics[width=\textwidth]{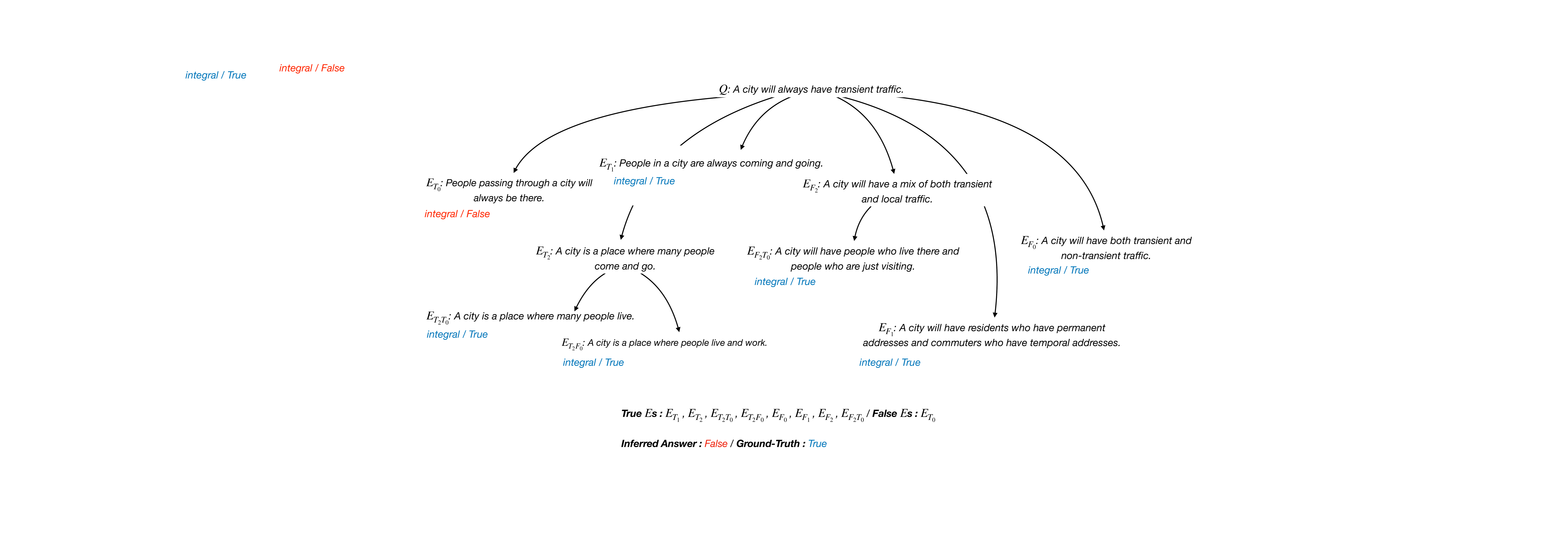}
    \caption{(continued) Example of incorrect inference by \method.}
\end{figure*}

\end{document}